\documentclass{article}

\usepackage[preprint]{ProbNum25} 

\usepackage[round]{natbib}

\usepackage{graphicx}
\graphicspath{{figures/}}
\usepackage{amsfonts,amsmath,amssymb}

\usepackage[capitalise,nameinlink]{cleveref}

\usepackage[disable]{todonotes}

\def\-{\text{-}}
\def\+{\text{+}}

\newcommand\given[1][]{\:#1\vert\:}

\probnumtitle{Bayesian autoregression to optimize temporal Matérn kernel Gaussian process hyperparameters}
\probnumauthors{%
\name{Wouter M. Kouw}%
\affiliation{Eindhoven University of Technology, Eindhoven, Netherlands}%
}

\probnumabstract{Gaussian processes are important models in the field of probabilistic numerics. We present a procedure for optimizing Matérn kernel temporal Gaussian processes with respect to the kernel covariance function's hyperparameters. It is based on casting the optimization problem as a recursive Bayesian estimation procedure for the parameters of an autoregressive model. We demonstrate that the proposed procedure outperforms maximizing the marginal likelihood as well as Hamiltonian Monte Carlo sampling, both in terms of runtime and ultimate root mean square error in Gaussian process regression.}

\begin{document}

\section{Introduction}
\label{sec:intro}
Temporal Gaussian processes are powerful probabilistic models to inter- and extrapolate time-series \citep{sarkka2013spatiotemporal}. They are used to forecast weather patterns such as rainfall and solar irradiation, to analyze brain recordings, and to monitor the structural health of windmills \citep{sarkka2013spatiotemporal,salcedo2014prediction,rogers2020application}. Optimizing the hyperparameters that govern the shape of the kernel covariance function (e.g., length scales) is notoriously challenging: the objective function may have strong local optima and regions of divergence \citep{rasmussen2006gaussian,svensson2015marginalizing}. These can cause gradient-based techniques, such as maximizing the marginal likelihood, to under-perform or even fail. Here we present a procedure that estimates the optimal hyperparameters quickly and robustly.

\begin{figure}[htb]
    \centering
    \includegraphics[width=\linewidth]{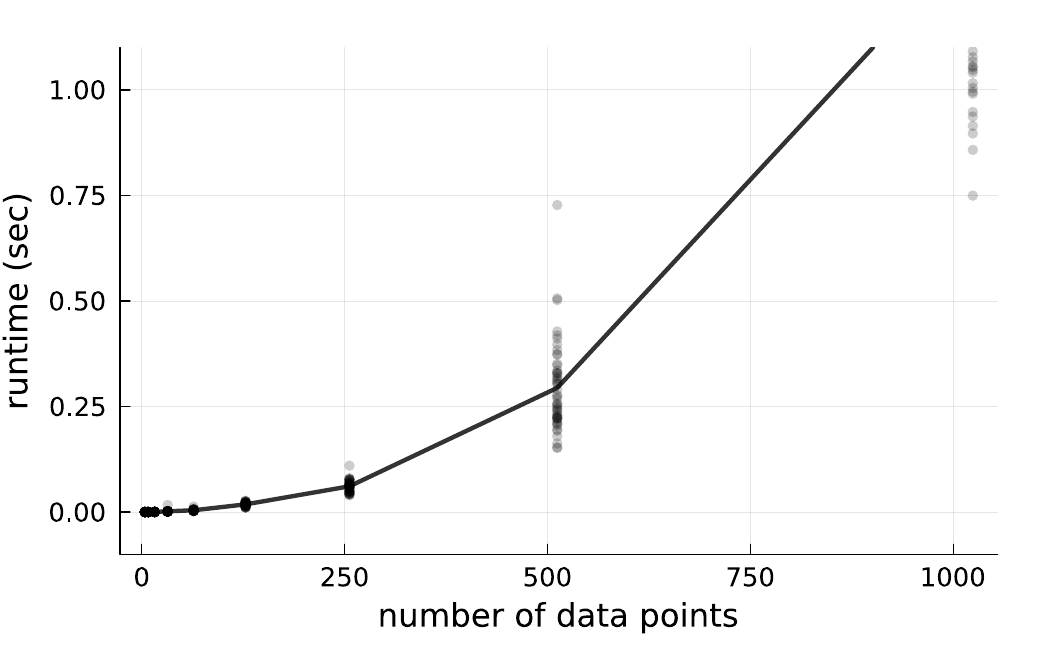}
    \caption{Simulation demonstrating the exponential increase in runtime for Gaussian process regression as the number of data points increases. The dots represent runtimes for  simulated experiments and the solid line represents their average.}
    \label{fig:problem_runtimes}
\end{figure}

Probabilistic numerics quantifies and tracks uncertainty as it propagates through a mathematical model or computational procedure \citep{hennig2022probabilistic}.  Traditional numerical methods are often limited to point estimates and are sensitive to noise or unexpected variations in measured signals. By incorporating sources of uncertainty explicitly, probabilistic numerical methods can adapt to perturbations and search spaces more effectively. The best example of this is probably Bayesian optimization, where instead of pursuing a gradient myopically, the optimizer infers the best possible next trial based on a balance between minimizing uncertainty and reaching the extremum \citep{garnett2023bayesian,hennig2022probabilistic}. Viewing optimization as an inference problem is rapidly leading to advances in speed and robustness \citep{zhilinskas1975single,hennig2013quasi,mahsereci2017probabilistic,mcleod2018optimization}. 
We focus on optimizing hyperparameters of Gaussian processes. The most well-studied procedure for this is maximization of the marginal likelihood \citep{ying1991asymptotic,ying1993maximum,karvonen2019asymptotics}. This is computationally expensive as fitting a Gaussian process is already expensive in the number of data points, but now the fit procedure must be repeated for every trial of the hyperparameters (Sec.~\ref{sec:problem}, Figure~\ref{fig:problem_runtimes}). Pre-conditioning techniques can accelerate this procedure \citep{wenger2022preconditioning}, but we believe further progress can be made by inferring the hyperparameters instead of optimizing them. Markov Chain Monte Carlo methods have produced robust procedures \citep{murray2010elliptical,svensson2015marginalizing}. We explore Bayesian filtering techniques with analytic solutions in order to improve in terms of speed. This is achieved for Matérn kernel temporal Gaussian process by casting them to linear stochastic differential equations, discretizing them to autoregressive processes and performing recursive Bayesian estimation on their parameters. Maximum a posteriori estimates of the autoregressive coefficients and noise precision parameter are reverted to kernel hyperparameters.
To be specific, our contributions consists of:
\begin{itemize}
    \item We formulate the temporal Gaussian process as an autoregressive difference equation using higher-order finite difference techniques.
    \item We propose a recursive Bayesian estimation procedure for kernel hyperparameters that relies on reverting a variable substitution.
\end{itemize}
The proposed procedure is validated on both simulated and real data, and compared against state-of-the-art methods.

\section{Problem Statement} \label{sec:problem}
Let $f : \mathbb{R}_{+} \rightarrow \mathbb{R}$ be a latent scalar function over time and $y_k \in \mathbb{R}$ be an observation of the function at time $t_k \in \mathbb{R}_{+}$ for positive real numbers $\mathbb{R}_{+}$. Observations are assumed to be noise-free, which implies our likelihood has the form of a Dirac delta function:
\begin{align} \label{eq:regression-likelihood}
    p(y_k \given f, t_k) = \delta(y_k - f(t_k)) \, .
\end{align}
Our prior distribution for the unknown function $f$ is a zero-mean Gaussian process:
\begin{align} \label{eq:GP-prior}
    p&\left(f \given t; \psi \right) = \mathcal{GP}\Big(f(t) \given 0, \kappa_{\psi}(t,t')\Big) \, ,
\end{align} 
with kernel covariance function $\kappa(\cdot)$ and hyperparameters $\psi$.
We focus on the Matérn class of stationary\footnote{Stationary covariance functions are only a function of a difference in time $t-t'$ \citep{rasmussen2006gaussian}.} covariance functions
\begin{align} \label{eq:matern}
\kappa^{\nu}_{\psi}(t,t') = \sigma^2 \frac{2^{1-\nu}}{\Gamma(\nu)} \Big(\frac{\sqrt{2\nu}}{l} |t \!- \! t'| \Big)^\nu \! B_\nu\Big(\frac{\sqrt{2\nu}}{l} |t \! - \! t'| \Big) \, ,
\end{align}
in which $\nu$ is a smoothness, $\sigma$ a magnitude, and $l$ a length scale hyperparameter \citep{rasmussen2006gaussian}. $\Gamma(\cdot)$ is a gamma function and $B_\nu(\cdot)$ is a modified Bessel function of the second kind. 
We restrict the smoothness parameter to half-integers, i.e., $\nu = m - 1/2$ for positive integers $m=1,2, \dots$. We will consider specific choices of $m$ (mainly $1$ and $2$) and let $\psi = (\sigma,l)$.

Suppose we have a data set of $N$ observations. Let $\mathbf{t} = \begin{bmatrix} t_1 \ \dots t_N \end{bmatrix}^{\intercal}$ and $\mathbf{y} = \begin{bmatrix} y_1 \ \dots \ y_N \end{bmatrix}^{\intercal}$.
Most commonly, the kernel hyperparameters are obtained by maximizing the marginal likelihood:
\begin{align} \label{eq:max-marginallik}
    \psi^{*} = \underset{\psi \in \Psi}{\arg \max} \ \ p(\mathbf{y} \given \mathbf{t} ; \psi)  \, ,  
\end{align}
The marginal likelihood is formed by integrating the likelihood (Eq.~\ref{eq:regression-likelihood}) over the Gaussian process prior distribution (Eq.~\ref{eq:GP-prior}) \citep{rasmussen2006gaussian}:
\begin{align}
    p(\mathbf{y} \given \mathbf{t}; \psi) &= \! \int \!p(\mathbf{y} \given f, \mathbf{t}) \, p(f \given \mathbf{t}; \psi) \ \mathrm{d}f \\
    &= (2\pi)^{-1/2} |K_\psi|^{-1/2} \exp \big(\! - \! \frac{1}{2} \mathbf{y}^{\intercal} K_\psi^{-1} \mathbf{y} \big) .
\end{align}
where $K_\psi$ is the Gram matrix of the kernel covariance function applied to all pairwise combinations of $t$.

The problem is that the maximization procedure scales poorly with the amount of data: in gradient-based optimization, a new $N \times N$ Gram matrix $K_\psi$ must be inverted for each iteration. It is thus $O(LN^3)$ where $L$ is the maximum number of gradient-based update steps. Figure \ref{fig:problem_runtimes} visualizes the runtime of this hyperparameter optimization procedure\footnote{Experimental details: we sampled 50 realizations of a Mat{\'e}rn-1/2 kernel Gaussian process over time steps of size $\Delta = 0.1$ with random sinusoidal outputs (frequency sampled from a uniform distribution over the interval $(0,2)$ and phase sampled uniformly from $(0,\pi)$) for $N = 2\char`\^\{2, \dots, 10\}$. We timed the marginal likelihood maximization procedure for each realization.} as a function of the number of data points. Note that the exponential nature.

\section{Solution Procedure}

Our hyperparameter optimization procedure consists of casting the Gaussian process to a linear stochastic differential equation, discretizing it in time using a higher-order finite difference method, inferring a posterior distribution over autoregressive model parameters and converting the maximum a posteriori estimates back to kernel hyperparameters. 

\subsection{Stochastic Differential Equations}
Temporal Gaussian processes with stationary Matérn kernel covariance functions have an equivalent representation as scalar linear stochastic differential equations \citep{hartikainen2010kalman,sarkka2013spatiotemporal}. 
For the sake of completeness, we shall restate this result briefly here. 

Consider an $m$-th order differential equation with coefficients $a_n$ for $n = 0, \dots, m-1$ driven by a white noise process $w(t)$ with spectral density $\varsigma^2$:
\begin{align} \label{eq:timedom}
\frac{d^m f(t)}{d t^m} + \sum_{n=0}^{m-1} a_n \frac{d^n f(t)}{dt^n} = w(t) \, .
\end{align}
Taking the Fourier transform on both sides gives
\begin{align} \label{eq:freqdom}
(i\omega)^m  F(\omega) + \sum_{n=0}^{m-1} a_n (i\omega)^n F(\omega) = W(\omega) \, ,
\end{align}
for frequency $\omega$. Note that the process in the frequency domain can be generated by a system with input $W(\omega)$ and a transfer function $H(\omega) = F(\omega) / W(\omega)$ that contains all the coefficients $a_n$ \citep{glad2018control}. 
The equivalent representation is based on the fact that the power spectral density of the process depends on the transfer function \citep{oppenheim1997signals}:
\begin{align} \label{eq:psd}
    S_F(\omega) = |H(\omega)|^2 S_W(\omega) = H(i\omega) \varsigma^2 H(-i\omega)\, .
\end{align}
Through the Wiener–Khinchin theorem, we know that the power spectral density of the Matérn kernel covariance function (Eq.~\ref{eq:matern}) is
\begin{align} \label{eq:psd-matern}
    S(\omega) = \sigma^2 \frac{ 2 \pi^{1/2} \Gamma(\nu \! + \! 1/2)}{\Gamma(\nu)} \lambda^{2\nu} (\lambda^2 + \omega^2)^{-(\nu + 1/2)} \, ,
\end{align}
where $\lambda = \sqrt{2\nu}/l$ \citep{rasmussen2006gaussian}.  So the coefficients $a_n$ and spectral density $\varsigma^2$ can be identified by matching Eq.~\ref{eq:psd-matern} and Eq.~\ref{eq:psd}.
The rational polynomial in $\omega^2$ can be factorized into two rational polynomials in $\omega$,
\begin{align}
    (\lambda^2 \+ \omega^2)^{-(\nu \+ \frac{1}{2})} = \underbrace{(\lambda \+ i\omega)^{-(\nu \+ \frac{1}{2})}}_{H(i\omega)} \underbrace{(\lambda - i\omega)^{-(\nu \+ \frac{1}{2})}}_{H(-i\omega)} ,
\end{align}
which are recognized as the two transfer functions in Eq.~\ref{eq:psd}.
The first function $H(i\omega) = (\lambda + i\omega)^{-(\nu + 1/2)}$ has its poles in the upper plane and will generate a stable stochastic process \citep{hartikainen2010kalman}. 
The binomial theorem generates the coefficients of its characteristic polynomial, which form the coefficients $a_n$ of Eq.~\ref{eq:freqdom}:
\begin{align} \label{eq:alpha_j}
    a_n = \binom{m}{n} \lambda^{m-n} \, .
\end{align}
The remaining terms become the spectral density of the white noise process, i.e.,
\begin{align}
    \varsigma^2 = \sigma^2 \lambda^{2\nu} \, 2 \pi^{1/2} \, \Gamma(\nu  +  1/2) / \Gamma(\nu) \, . \label{eq:psd-wiener}
\end{align}


\paragraph{Examples} We will use two running examples throughout the paper to illustrate steps: the $\nu = 1/2$ (i.e., $m=1$) and $\nu = 3/2$ (i.e., $m = 2$) cases. Firstly, for $\nu = 1/2$, we have:
\begin{align}
H(i\omega) = (\lambda + i\omega)^{-1}  \, .
\end{align}
This gives a first-order stochastic process, 
\begin{align}
    (i\omega) F(\omega) + \lambda F(\omega) = W(\omega)  \, ,
\end{align}
whose time domain form is
\begin{align}
     \frac{d f(t)}{d t} + \lambda f(t) = w(t) \, .    \label{eq:nu12-timedom}
\end{align}
So we have only one coefficient, that is $a_0 = \lambda$. For the white noise, we have:
\begin{align}
    \varsigma^2 = \sigma^2 \lambda \, 2 \pi^{1/2} \, \Gamma(1) / \Gamma(1/2) = 2 \sigma^2 \lambda  \, .
\end{align}

For the $\nu = 3/2$ case, the transfer function is 
\begin{align}
H(i\omega) = (\lambda + i\omega)^{-2} = (\lambda^2 + 2 \lambda i\omega + (i\omega)^2)^{-1} \, .
\end{align}
This gives an order $m=2$ stochastic process, 
\begin{align}
    (i\omega)^2 F(\omega) + 2 \lambda (i\omega) F(\omega) + \lambda^2 F(\omega) = W(\omega)  \, ,
\end{align}
whose time domain form is
\begin{align}
    \frac{d^2 f(t)}{d t^2} + 2 \lambda \frac{d f(t)}{d t} +  \lambda^2 f(t)  = w(t) \, .    
\end{align}
So, in this case $a_1 = 2\lambda$ and $a_0 = \lambda^2$. For the white noise, we have:
\begin{align}
    \varsigma^2 = \sigma^2 \lambda^3 \, 2 \pi^{1/2} \, \Gamma(2) / \Gamma(3/2) = 4 \sigma^2 \lambda^3  \, .
\end{align}

\subsection{Time Discretization}
In seminal earlier work \citep{hartikainen2010kalman,sarkka2012infinite,sarkka2013spatiotemporal}, Eq.~\ref{eq:timedom} is represented as a Markov process with state vector $x(t) = \begin{bmatrix} f(t) & d f(t)/dt & \dots & dt^{m-1} f(t)/dt^{m-1} \end{bmatrix}$,
\begin{align} \label{eq:tgp-diffeq}
    \frac{d x(t)}{dt} = F x(t) +  L w(t) \, .
\end{align}
The transition matrix and noise matrices are:
\begin{align}
    F  =  \begin{bmatrix} 0 & 1 &  \dots & 0 \\ \vdots & 0 & 1 &  \vdots \\ \vdots & \vdots & \ddots & \vdots \\ -a_0 & -a_1 &  \dots & - a_{m\-1} \end{bmatrix} \, , \ L = \begin{bmatrix} 0 \\ \vdots \\ 1 \end{bmatrix} .
\end{align}
This is treated as a state-space model, and Bayesian filtering techniques are used to estimate the value of $f(t)$ at specific times. We will explore a different direction, using higher-order finite difference techniques to cast Eq.~\ref{eq:timedom} as a discrete-time autoregressive process.

Consider an explicit forward finite difference approximation to an order-$m$ derivative \citep{hamming2012introduction}:
\begin{align}
    \frac{d^{m} f(t)}{dt^{m}} \approx \frac{1}{\Delta^m} \sum_{n=0}^{m} (-1)^{m-n} \binom{m}{n} f(t + n \Delta) \, ,
\end{align}
where $\Delta$ is a fixed step size, i.e., $\Delta = t_k - t_{k\-1}$ for all $k$. Let $f_k = f(t_k), f_{k\+1} = f(t_k + \Delta), f_{k\+2} = f(t_k + 2\Delta)$ and so on.
Applying the forward finite difference approximation to each of the derivatives in Eq.~\ref{eq:timedom} produces
\begin{align}
    \sum_{n=0}^{m} a_n \frac{d^n f(t)}{d t^n} \approx \sum_{n=0}^{m} \frac{a_n}{\Delta^n} \sum_{j=0}^{n} (-1)^{n-j} \binom{n}{j} f_{k\+j} \, .
\end{align}
In a time step of size $\Delta$, $w(t)$ contributes to the process with an increment drawn from a zero-mean Gaussian $w_k$ with variance $\varsigma^2 \Delta$ \citep{sarkka2019applied}; 
\begin{align}
    \int_{0}^{\Delta} w(s) ds = w_k \sim \mathcal{N}\left(0, \varsigma^2 \Delta \right) \, .
\end{align}
We have thus discretized the stochastic differential equation in Eq.~\ref{eq:timedom} to:
\begin{align} \label{eq:sde-disc}
    \sum_{n=0}^{m} \frac{a_n}{\Delta^n} \sum_{j=0}^{n} (-1)^{n-j} \binom{n}{j} f_{k+j} = w_k \, .
\end{align}
We would like to isolate the most forward function step on one side of the equation;
\begin{align} \label{eq:maxforward}
    &\sum_{n=0}^{m} \frac{a_n}{\Delta^n} \sum_{j=0}^{n} (-1)^{n-j} \binom{n}{j} f_{k\+j} = \frac{a_m}{\Delta^m} f_{k\+m} + \\
    &\  \sum_{n=0}^{m-1} \frac{a_m}{\Delta^m} (\-1)^{m\-n} \binom{m}{n} f_{k\+n} \+ \sum_{j=0}^{n} \frac{a_j}{\Delta^j} (\-1)^{n\-j} \binom{n}{j} f_{k\+j} . \nonumber 
\end{align}
Note that, due to the nature of the transfer function, the coefficient belonging to the highest-order derivative, $a_m$, will always be $1$ (Eq.~\ref{eq:timedom}). That means the leading coefficient is $1/\Delta^{m}$. Multiplying by $\Delta^m$ and using Eq.~\ref{eq:maxforward}, Eq.~\ref{eq:sde-disc} becomes:
\begin{align} 
    &f_{k+m} = -\sum_{n=0}^{m-1} (-1)^{m-n} \binom{m}{n} f_{k\+n} \label{eq:autoregressive-diffeq1} \\ 
    & \qquad \qquad \ - \sum_{j=0}^{n} a_j\Delta^{m-j} (-1)^{n-j} \binom{n}{j} f_{k\+j}  +  \Delta^{m} w_k  \, . \nonumber
\end{align}
Under this time discretization scheme, the stochastic differential equation representation of the temporal Gaussian process becomes an autoregressive process (Sec.~\ref{sec:modelspec}).

\paragraph{Examples} For the $m=1$ case, the forward finite difference is familiar
\begin{align}
    \frac{d f(t)}{dt} &\approx \frac{f_{k\+1} - f_k}{\Delta} \, .
\end{align}
Under this scheme, the weighted sum of derivatives is approximately:
\begin{align}
    a_1 \frac{d f(t)}{dt} + a_0 f(t) &\approx a_1 \frac{f_{k\+1} - f_k}{\Delta} + a_0 f_k \\
    &= \frac{a_1}{\Delta} f_{k\+1} + (a_0 - \frac{a_1}{\Delta}) f_k \, .
\end{align}
Equating this to the discretized noise and moving terms around yields:
\begin{align}
    \frac{a_1}{\Delta} &f_{k\+1} + (a_0 - \frac{a_1}{\Delta}) f_k = w_k \\
    &f_{k\+1} = (1 - a_0 \Delta)  f_k + \Delta w_k \, . \label{eq:nu12-ar}
\end{align}

For the $m=2$ case, the forward finite difference is:
\begin{align}
    \frac{d^2 f(t)}{dt^2} &\approx \frac{f_{k\+2} - 2 f_{k\+1} + f_k}{\Delta^2} \, .
\end{align} 
Incorporating this into the weighted sum of derivatives yields:
\begin{align}
    &a_2 \frac{d^2 f(t)}{d t^2} 
    + a_1 \frac{d f(t)}{d t} + a_0 f(t) \nonumber \\
    &\approx \quad  \frac{a_2}{\Delta^2} \left( \binom{2}{0} f_k - \binom{2}{1} f_{k\+1} \+ \binom{2}{2} f_{k\+2} \right) \nonumber \\
    &\quad + \frac{a_1}{\Delta^1} \left( \binom{1}{0} f_k - \binom{1}{1} f_{k\+1} \+ \binom{1}{2} f_{k\+2} \right) \nonumber \\
    &\quad + \frac{a_0}{\Delta^0} \left( \binom{0}{0} f_k - \binom{0}{1} f_{k\+1} \+ \binom{0}{2} f_{k\+2} \right) \\
    &= \frac{a_2}{\Delta^2} \big(f_k \! - \! 2 f_{k\+1} \+ f_{k\+2}) \+ \frac{a_1}{\Delta} \big(f_{k} \! - \! f_{k\+1} \big) \+ a_0 f_k \\
    &= \frac{a_2}{\Delta^2} f_{k\+2} - (\frac{2a_2}{\Delta^2} \+ \frac{a_1}{\Delta} ) f_{k\+1} \+ (\frac{a_2}{\Delta^2} \+  \frac{a_1}{\Delta} \+ a_0) f_k .
\end{align}
Equating this to the discretized noise, setting $a_2 = 1$, and moving terms around yields:
\begin{align}
    &\frac{a_2}{\Delta^2} f_{k\+2} - (\frac{2a_2}{\Delta^2} + \frac{a_1}{\Delta} ) f_{k\+1} + (\frac{a_2}{\Delta^2} +  \frac{a_1}{\Delta} + a_0) f_k = w_k \\
    &f_{k\+2} =  (2 + a_1 \Delta ) f_{k\+1} - (1 +  a_1 \Delta + a_0 \Delta^2) f_k + \Delta^2 w_k \, .
\end{align}

\subsection{Probabilistic Model Specification} \label{sec:modelspec}
In order to specify a probabilistic model and infer the kernel hyperparameters $\lambda, \sigma$, we perform a number of variable substitutions.
First, we combine the terms involving $f$ on the right-hand side of Eq.~\ref{eq:autoregressive-diffeq1} into:
\begin{align} \label{eq:term-combination}
    - \sum_{n=0}^{m-1} (-1)^{m\-n} \binom{m}{n} f_{k\+n} \! - \! \sum_{j=0}^{n} & a_j\Delta^{m\-j} (\-1)^{n\-j} \binom{n}{j} f_{k\+j} \nonumber \\
    &= \sum_{n=0}^{m-1} \theta_n f_{k\+n}  \, ,
\end{align}
where the substitutions are
\begin{align} \label{eq:arcoeff-sub}
    \theta_n \! = \! (-1)^{m\-n\+1} \binom{m}{n} - \sum_{j=0}^{m\-1} a_n \Delta^{m\-n} (-1)^{j\-n} \binom{j}{n} .
\end{align}
Plugging Eq.~\ref{eq:term-combination} into Eq.~\ref{eq:autoregressive-diffeq1} reveals the autoregressive nature of the higher-order finite difference approximation of the stochastic differential equation:
\begin{align}
    f_{k+m} = \sum_{n=0}^{m-1} \theta_n f_{k\+n}  + \Delta^m w_k \, .
\end{align}
We shall therefore refer to $\theta_n$ as the autoregressive coefficients.

To substitute for the scale of the Gaussian noise contribution, we first incorporate the $\Delta^m$ scaling factor into the variance of $w_k$. Let $\bar{w}_k = \Delta^m w_k$ such that $\bar{w}_k \sim \mathcal{N}(0, \Delta^{2m+1} \varsigma^2)$. Then we define a noise precision parameter $\tau$ as:
\begin{align} \label{eq:noise-sub}
    \tau &= 1/(\Delta^{2m+1} \varsigma^2) \\
    &= 1/\big(\Delta^{2m+1} \sigma^2 \lambda^{2\nu} \, 2 \pi^{1/2} \, \Gamma(\nu  +  1/2) / \Gamma(\nu) \big) \, .
\end{align}
The kernel hyperparameters are part of the autoregressive coefficients and noise precision parameters. We can now specify a standard probabilistic autoregressive model, pose suitable prior distributions and infer posterior distributions. Reverting the variable substitutions using the maximum a posteriori estimates will give us estimates of the optimal kernel hyperparameters $\psi^{*}$.

\paragraph{Examples} For $\nu = 1/2$ (i.e., $m=1$), we combine Eqs.~\ref{eq:nu12-ar} and \ref{eq:nu12-timedom} to find:
\begin{align} \label{eq:sub-ar1}
    \theta_0 &= 1 - a_0 \Delta  = 1 - \lambda \Delta \\
    \tau &= (\varsigma^2 \Delta^3)^{-1} = 1/(2 \sigma^2 \lambda \Delta^3) \, . 
\end{align}
In the $\nu = 3/2$ ($m=2$) case, the substitutions are:
\begin{align} 
    \theta_0 &= 2 + a_1 \Delta = 2 + 2\lambda \Delta \label{eq:sub-ar2-1} \\
    \theta_1 &= - (1 \+ a_1 \Delta \+ a_0 \Delta^2) = - 1 - 2\lambda \Delta - \lambda^2 \Delta^2 \label{eq:sub-ar2-2} \\
    \tau &= (\varsigma^2 \Delta^5)^{-1} = 1/( 4\sigma^2 \lambda^3 \Delta^5) \label{eq:sub-ar2-3} \, . 
\end{align}

\paragraph{Likelihood Function}
Given the variable substitution above, we can define a likelihood function. Consider a buffer of previous observations\footnote{For $k < m-1$, note that the buffer is initialized with zeros and is filled with observations as they arrive.}
\begin{align}
    \bar{y}_k = \begin{bmatrix} y_k & y_{k-1} & \dots & y_{k-m-1} \end{bmatrix} \, .
\end{align}
Then we may write
\begin{align}
    p(y_{k+1} \given \bar{y}_k, \theta, \tau) = \mathcal{N}(y_{k+1} \given \theta^{\intercal} \bar{y}_k, \tau^{-1}) \, ,
\end{align}
for unknown coefficients $\theta = \begin{bmatrix} \theta_0 \ \dots \ \theta_{m-1} \end{bmatrix}^{\intercal}$ and noise precision parameter $\tau$.

\paragraph{Prior Distribution} We specify a joint prior distribution on the autoregression coefficient $\theta$ and the noise precision $\tau$. Specifically, a compound multivariate Gaussian and univariate Gamma distribution:
\begin{align}
    p(\theta, \tau) &= \mathcal{NG}(\theta, \tau \given \mu, \Lambda, \alpha, \beta) \\
    &= \mathcal{N}(\theta \given \mu, (\tau \Lambda)^{-1}) \, \mathcal{G}(\tau \given \alpha, \beta) \, ,
\end{align}
with mean $\mu$, precision matrix $\Lambda$, shape parameter $\alpha$ and rate parameter $\beta$. Their density functions are:
\begin{align}
    \mathcal{N}(\theta | \mu, (\tau \Lambda)^{\-1}) &= \! \frac{|\tau \Lambda|^{1/2}}{(2\pi)^{m/2}} \exp( -  \frac{\tau}{2} (\theta \- \mu )^{\intercal} \Lambda (\theta \-  \mu ) )   \\
    \mathcal{G}(\tau \given \alpha, \beta) &= \frac{\beta^{\alpha}}{\Gamma(\alpha)} \tau^{\alpha-1} \exp(-\beta \tau) \, .
\end{align}
This choice of priors is conjugate to the autoregressive likelihood \citep{shaarawy2008bayesian,kouw2023information}.

\subsection{Bayesian Inference Procedure}
We will adopt a Bayesian filtering procedure to obtain a posterior distribution over parameters \citep{sarkka2023bayesian}:
\begin{align}
    p(\theta, \tau \given y_{1:k\+1}) = \frac{p(y_{k\+1} \given \bar{y}_k, \theta, \tau)}{p(y_{k\+1} \given y_{1:k})} p(\theta, \tau \given y_{1:k}) .
\end{align}
As the prior distribution is conjugate to the autoregressive likelihood, it yields an exact posterior distribution
\begin{align}
    p(\theta, \tau | y_{1:k\+1}) \! = \! \mathcal{NG}\big(\theta,\tau | \mu_{k\+1}, \Lambda_{k\+1}, \alpha_{k\+1}, \beta_{k\+1} \big) ,
\end{align}
with parameters:
\begin{align} \label{eq:truepost_params}
    &\mu_{k\+1} = \big(\bar{y}_k \bar{y}_k^{\intercal}  +  \Lambda_{k} \big)^{-1}\big(\bar{y}_k y_{k\+1}  +  \Lambda_{k} \mu_{k} \big)  \\[1.0ex]
    &\Lambda_{k\+1} = \bar{y}_k \bar{y}_k^{\intercal}  +  \Lambda_{k} \\[0.5ex]
    &\alpha_{k\+1} = \alpha_{k} + \frac{1}{2}  \\[0.5ex]
    &\beta_{k\+1} = \beta_{k} \+  \frac{1}{2}\big( y_{k\+1}^2 \! - \!  \mu_{k\+1}^{\intercal} \Lambda_{k\+1} \mu_{k\+1} \+ \mu_{k}^{\intercal}\Lambda_{k} \mu_{k} \big) .
\end{align}
This is a recursive solution, which means that we start with a set of initial points $(\mu_0,\Lambda_0,\alpha_0,\beta_0)$ and that each update aims to produce a better estimate. Whether there is a strict improvement is an open question. Note that these parameters can be reverted at any time to produce an estimate of the kernel hyperparameters. In that sense, we have a running solution, a valuable property when the computational budget is limited.

\subsection{Reverting the Substitution}

The variable substitutions in Equations \ref{eq:arcoeff-sub} and \ref{eq:noise-sub} are a system of polynomial equations. Reverting the substitution means finding values for $\lambda$ and $\sigma$ for which every polynomial equation is equal to the substituted variables $\theta, \tau$. In other words, we must solve the system of polynomial equations \citep{sturmfels2002solving}. For $m \geq 2$, the system is overdetermined, i.e., there are more equations than unknowns, and it is non-homogeneous (i.e., it has non-zero constants), which means it will probably have no real solutions \citep{chen2015homotopy}. If there are none, the substituted variables cannot be reverted exactly. However, we can still look for approximate reversions that may prove to perform well. Here we opt for a nonlinear least-squares approach. We construct an objective function that consists of the sum of squared deviations of every polynomial from the MAP estimates of the autoregressive parameters, $\mu_n$ and $(\alpha - 1)/\beta$. These are:
\begin{align}
    g_n(\psi) = \Big(\mu_n - \theta_n \Big)^2 \, , \quad 
     g_m(\psi) = \Big(\frac{\alpha - 1}{\beta} - \tau \Big)^2 \, ,
\end{align}
for $n = 0, \dots,  m-1$, $\theta_n$ as defined in Eq.~\ref{eq:arcoeff-sub} and $\tau$ in Eq.~\ref{eq:noise-sub}.
The objective is minimized with respect to $\psi = (\lambda, \sigma)$ with the constraint that both should be positive real-valued numbers $\Psi \in \mathbb{R}_{+}^2$;
\begin{align} \label{eq:approx-reverse-nonlinls}
    \psi^{*} = \underset{\psi \in \Psi}{\arg \min} \ \sum_{i=0}^m \, g_i(\psi) \, . 
\end{align}
Note that this optimization problem scales with the degree of the Mat{\'e}rn kernel $m$, and not the number of training data points $N$ (as is the case in the original marginal likelihood maximization problem of Eq.~\ref{eq:max-marginallik}).

\paragraph{Examples} For the $\nu=1/2$ ($m=1$) case, the system is linear and can be reverted exactly,
\begin{align}
    \mu = 1 - \lambda \Delta \quad  \implies \quad \lambda = \frac{1 - \mu}{\Delta} \, .
\end{align}
This result is plugged into the noise substitution,
\begin{align}
    \frac{\alpha - 1}{\beta} = 1/ \left( 2 \sigma^2 \lambda \Delta^3 \right) 
    = 1/\left(2 \sigma^2 (1 - \mu) \Delta^2 \right) \, ,
\end{align}
yielding the reversion
\begin{align}
    \sigma^2 = \frac{\beta}{2 (\alpha - 1) (1 - \mu) \Delta^2} \, .
\end{align}
Note that this reversion can be done at any time, i.e., for any $k = 1, \dots,  N$.

The $\nu = 3/2$ ($m = 2$) case is a system of polynomial equations. The application of homotopy continuation techniques reveals that it has no real solutions \citep{chen2015homotopy}. Plugging in the specific polynomials from Equations \ref{eq:sub-ar2-1}-\ref{eq:sub-ar2-3}, produces the following squared error terms for the nonlinear least-squares objective:
\begin{align}
    &g_0(\psi) = \Big(\mu_1 - (2 + 2\lambda \Delta)\Big)^2 \\
    &g_1(\psi) = \Big(\mu_2 - (- 1 - 2\lambda \Delta - \lambda^2 \Delta^2)\Big)^2 \\
    &g_2(\psi) = \Big(\frac{\alpha-1}{\beta} - 1/( 4\sigma^2 \lambda^3 \Delta^5)\Big)^2 \, .
\end{align}
Minimizing the sum of these functions with respect to $\psi$ produces approximate reversions of the autoregressive parameters to the kernel hyperparameters.

\section{Experiments}

We perform simulation experiments with randomly generated temporal Gaussian processes\footnote{For code and additional experimental details, see \href{https://github.com/biaslab/ProbNum2025-BARGPhparams}{github.com/biaslab/ProbNum2025-BARGPhparams}.}. The proposed probabilistic solution (referred to as BAR) is compared to the two most commonly used procedures: maximizing the marginal likelihood with respect to the kernel hyperparameters (Eq.~\ref{eq:max-marginallik}; referred to as MML) and Hamiltonian Monte Carlo sampling (referred to as HMC). For this, we employ the state-of-the-art toolbox GaussianProcesses.jl \citep{fairbrother2022gaussianprocesses}. The hyperparameters are expressed on a log-scale to allow for unconstrained optimization. The mean of the prior distribution is fixed to $0$'s, the initial points for MML are $1.0$, the priors for HMC are Exponential distributions, and the maximum number of iterations is limited to $1000$.
To solve the optimization problem in Eq.~\ref{eq:approx-reverse-nonlinls}, we use Optim.jl with a log-barrier function to enact the positivity constraints and L-BFGS as optimizer \citep{mogensen2018optim}.

\paragraph{Evaluation} To evaluate the found hyperparameters, we plug them into the standard predictive distribution formulation of a Mat{\'e}rn-kernel Gaussian process regression model. The degree of the Mat{\'e}rn kernel for testing will match the one for training. We express performance as the root mean square error (RMSE) between the mean vector of the predictive distribution and the observed data points.

\subsection{Simulations}

\paragraph{Matérn-1/2 Kernel} We generate 50 realizations of an isotropic Mat{\'e}rn-1/2 kernel Gaussian process as training set, and another 50 as test set. The system $\lambda$ was sampled from a Beta distribution with a shape parameter of $10$ and a rate parameter of $4$ and the system noise precision parameter $\tau$ was sampled from a Gamma distribution with shape parameter $10$ and rate parameter $1$. The time step size was $\Delta = 0.1$ and the outputs were generated from sinusoid with frequency sampled from a uniform distribution over the interval $(0,2)$ and phase sampled uniformly from $(0,\pi)$.
For the prior parameters of the autoregressive solution, we used $\mu_0 = 0.0$, $\Lambda_0 = 10^{-3}$, $\alpha_0 = 2$ and $\beta_0 = 0.1$. These should be considered as weakly informative, i.e., not flat but not concentrated on the system parameters either. 

\begin{figure}[htb]
    \centering
    \includegraphics[width=\linewidth]{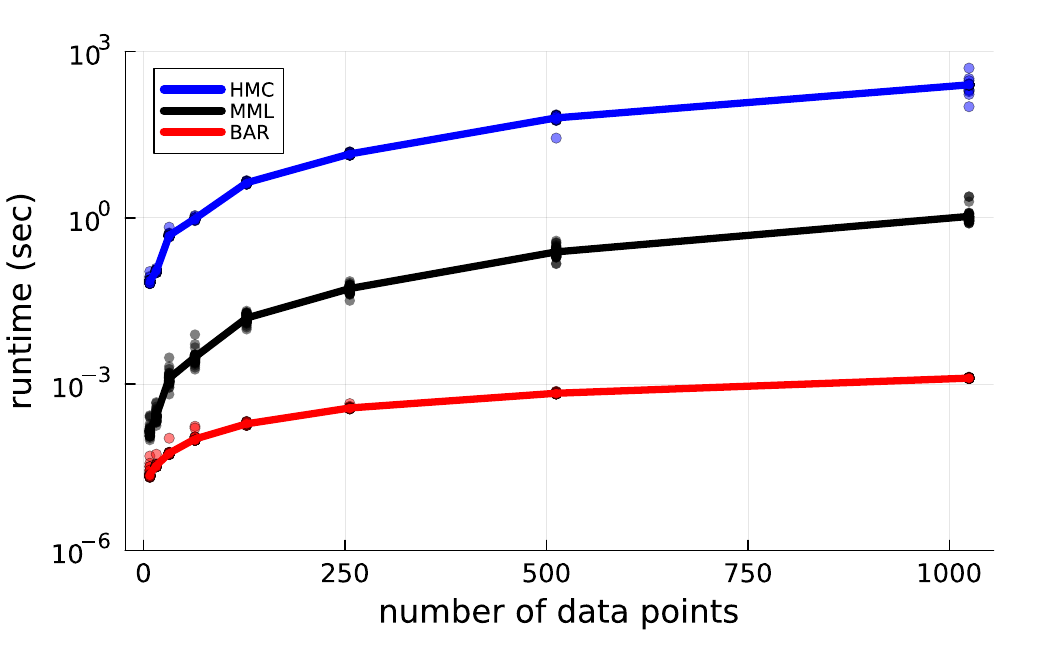}
    \includegraphics[width=\linewidth]{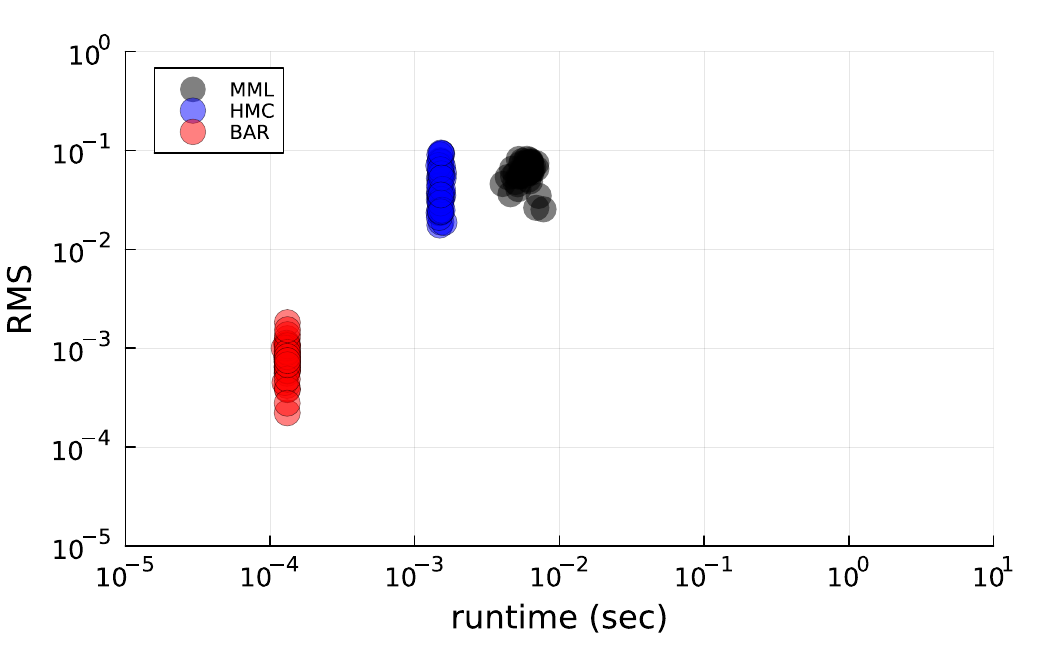}
    \caption{Mat{\'e}rn-1/2. Comparison of maximizing the marginal likelihood (MML), Hamiltonian Monte Carlo (HMC) versus Bayesian autoregression (BAR) for runtime (in seconds) as a function of the number of training data points (top) and root mean square error as a function of runtime (bottom; $N=100$).}
    \label{fig:experiment:AR1}
\end{figure}

In the first experiment, we time both procedures as a function of the number of data points $N$. Figure \ref{fig:experiment:AR1} (top) demonstrates that BAR is much faster and scales better than both MML and HMC.
In the second experiment, the training and test data set sizes are fixed to $N=100$ and we compare the ultimate root mean square error of the estimated hyperparameters as a function of the procedure's runtime. Figure \ref{fig:experiment:AR1} (bottom) shows that BAR dominates not just in runtime, but also in terms of RMSE. 

\begin{figure}[htb]
    \centering
    \includegraphics[width=\linewidth]{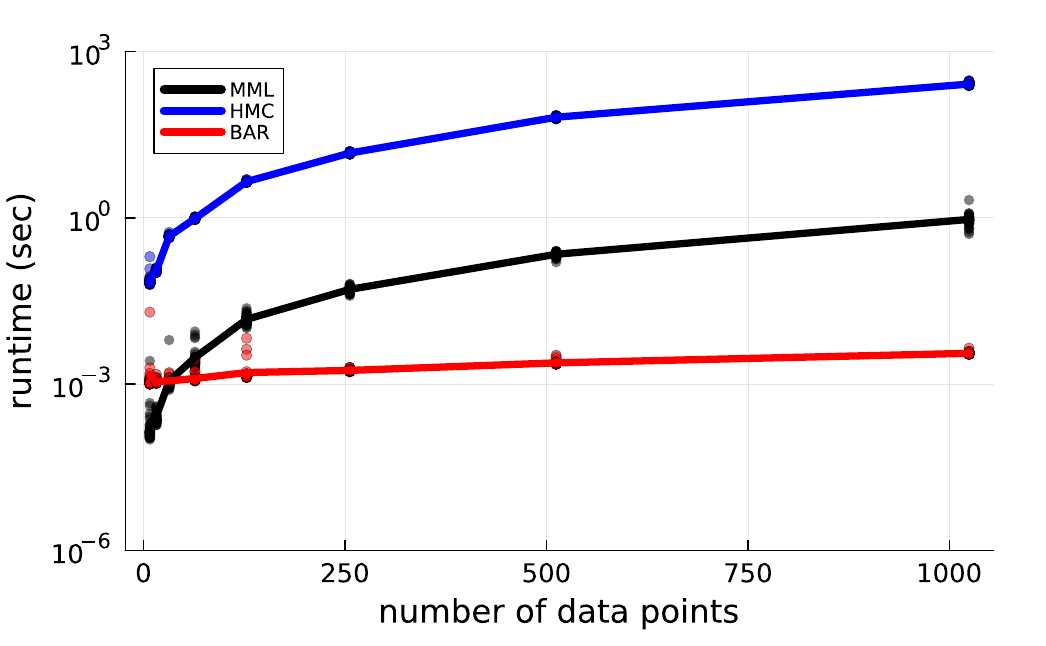}
    \includegraphics[width=\linewidth]{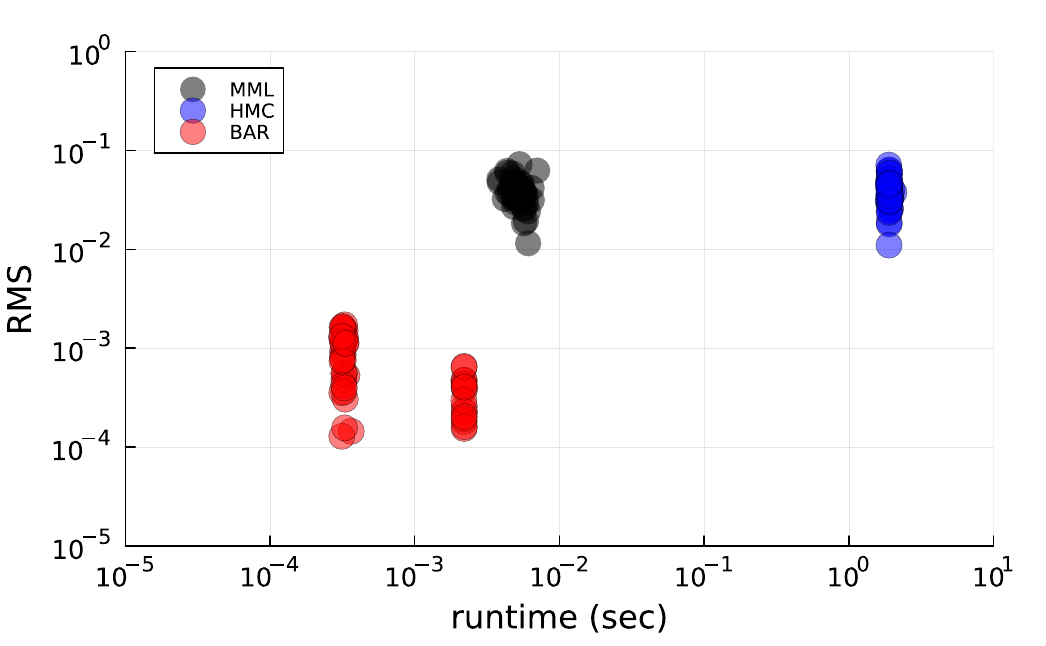}
    \caption{Mat{\'e}rn-3/2. Comparison of maximizing the marginal likelihood (MML), Hamiltonian Monte Carlo (HMC) and Bayesian autoregression (BAR) for runtime (in seconds) as a function of the number of training data points (top) and root mean square error as a function of runtime (bottom; $N=100$).}
    \label{fig:experiment:AR2}
\end{figure}

\paragraph{Matérn-3/2 Kernel}
We generate 10 realizations of an isotropic Mat{'e}rn-3/2 kernel Gaussian process as training set, and another 10 as test set. As before, the system $\lambda$ was sampled from a Beta distribution with a shape parameter of $10$ and a rate parameter of $4$ and the system noise precision parameter $\tau$ was sampled from a Gamma distribution with shape parameter $10$ and rate parameter $1$. The time step size was still $\Delta = 0.1$ and the outputs were also generated from sinusoid with frequency sampled from a uniform distribution over the interval $(0,2)$ and phase sampled uniformly from $(0,\pi)$.
Figure \ref{fig:experiment:AR2} (top) again shows runtime comparisons, demonstrating that, although BAR is slower than MML for $N \leq 16$, it scales much better than MML and HMC (factor $10^3$ improvement over MML and $10^6$ improvement over HMC).

Figure \ref{fig:experiment:AR1} (bottom) shows that BAR still outperforms MML and HMC in terms of RMSE by a wide margin.

\subsection{Real data}

\begin{figure}[htb]
    \includegraphics[width=.49\textwidth]{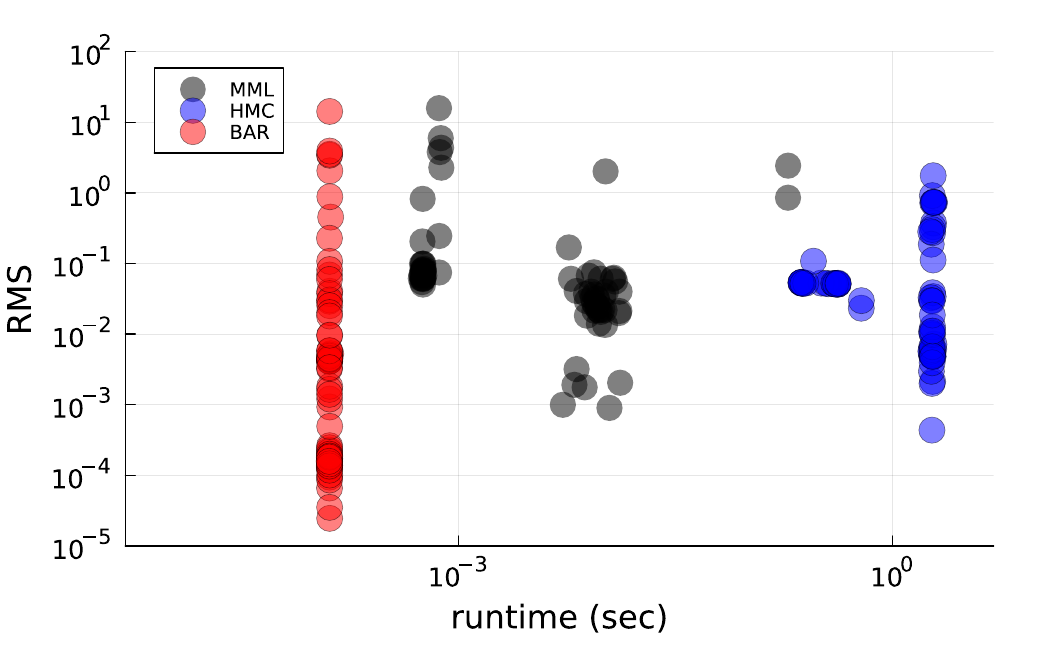} \\
    \includegraphics[width=.49\textwidth]{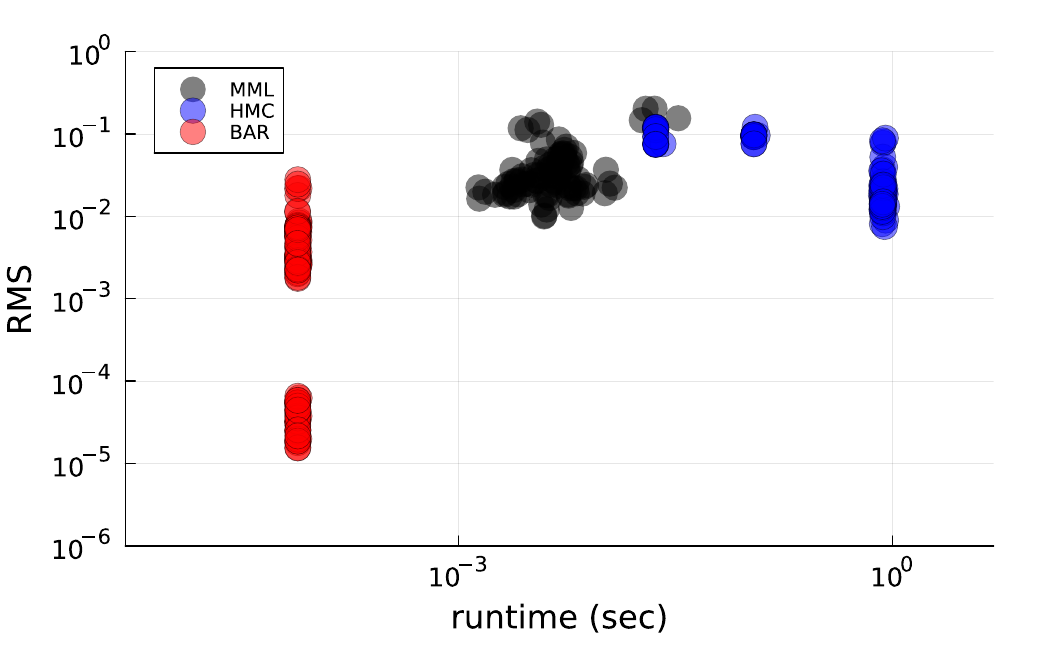}
    \caption{Comparison of BAR, HMC and MML for a Mat{\'e}rn-1/2 kernel Gaussian process in terms RMSE versus runtime. (Top) Room occupancy data set. (Bottom) Condition monitoring of hydraulic system data.}
    \label{fig:experiment:real}
\end{figure}

\paragraph{Room Occupancy} Various measurements were taken to estimate the number of occupants in a room, using non-intrusive devices that sense temperature, (infrared) light, CO2 and sound \citep{singh2018machine}. We subsampled the data to a measurement every $2$ minutes, leaving 2531 data points. We performed $100$ experiments where we sampled uniformly at random one of the 17 features and a starting point between $1$ and $2531$. We then created the training set from the first $N=100$ points after the start, and the test set from the $100$ after that. Figure \ref{fig:experiment:real} (top) shows the RMSE versus runtime comparison of the three methods. BAR dominates the other two in terms of runtime. On average, it performs better than MML and HMC but there are data sets for which it performs worse than MML or HMC.

\paragraph{Monitoring Hydraulics}
In this data set, a hydraulic test rig's condition was monitored over time using a range of sensors \citep{helwig2015condition}. We focused on temperature, vibration and cooling power, all read out once per second. The measurements are taken over multiple cycles that last 60 seconds ($N = 60$ for a single cycle). We sampled uniformly at random 1 of the 3 sensors and 2 of the 2205 cycles (1 for training and 1 for testing). Figure \ref{fig:experiment:real} plots the RMSE versus runtime for all three methods. BAR dominates in terms of runtime and outperforms the other methods in terms of RMSE in nearly all cases.

\section{Discussion}

For the Mat{\'e}rn-1/2 kernel, we effectively turned an optimization problem into a Bayesian filtering one. For higher order Mat{\'e}rn kernels, we have turned an optimization problem that scales on the order of the number of data points into one that scales on the order of the kernel. Although this will produce solutions faster, it remains to be analyzed how the approximate reversion relates to the optimal hyperparameters.
The proposed method is specific to Mat{\'e}rn kernel Gaussian processes and will not easily generalize to other kernel covariance functions. It may be possible to utilize it for the squared exponential kernel; it can be expressed as a stochastic differential equation but it is of infinite-order and would require a finite-order approximation \citep{hartikainen2010kalman,sarkka2013spatiotemporal}. In terms of experimental validation, it should be noted that there are many more methods for obtaining kernel hyperparameters for Gaussian processes (e.g., Bayesian optimization, sequential Monte Carlo), and more work is needed to fully characterize the position of the proposed approach in that landscape \citep{rasmussen2006gaussian,murray2010elliptical,svensson2015marginalizing}.

There are several points of improvement for the proposed solution. Firstly, we have yet to incorporate noisy observations. This could be achieved by incorporating the observation noise into the variable substitution in Section \ref{sec:modelspec}. However, if the observation noise precision parameter is unknown, then this would complicate the variable substitution reversion. Secondly, it is unclear what the effect of the approximation errors (high-order finite differences, numerical solution to system of polynomial equations) is on the Bayesian parameter estimates and on the subsequent kernel hyperparameter estimates; does it lead to bias? Thirdly, it remains to be studied whether a Bayesian autoregressive inference procedure produces consistent estimates of the kernel hyperparameters. The maximal marginal likelihood estimator is strongly consistent and asymptotically normal \citep{ying1993maximum}, but the approximation errors inherent to our proposed procedure prevent a similar analysis.

\section{Conclusion}
We presented a Bayesian inference procedure for obtaining the hyperparameters of Matérn kernel temporal Gaussian processes, models important for time series analysis and forecasting. The solution was based on casting the stochastic differential equation representation of the temporal Gaussian process as an autoregressive difference equation and then performing recursive Bayesian estimation of the autoregressive coefficients and noise precision parameter. The proposed procedure was faster than state-of-the-art methods and produced estimates that led to improved predictive performance in subsequent Gaussian process regression.

\bibliography{references}

\end{document}